\documentclass[11pt]{article}

\usepackage[preprint]{acl}

\usepackage{times}
\usepackage{latexsym}

\usepackage[T1]{fontenc}

\usepackage[utf8]{inputenc}

\usepackage{microtype}

\usepackage{inconsolata}

\usepackage{graphicx}

\usepackage{hyperref}
\usepackage{url}

\usepackage{booktabs}        
\usepackage{multirow}        
\usepackage{multicol}       
\usepackage{array}           
\usepackage{tabularx}      
\usepackage{graphicx}       
\usepackage[table]{xcolor}      
\usepackage{tikz}           
\usepackage{pgfplots}
\usepackage{algorithm}
\usepackage{algorithmic}
\usepackage{enumitem} 
\usepackage{amsmath}
\usepackage{subcaption}
\usepackage{amsfonts}
\usepackage{amssymb}
\usepackage{xspace}
\usepackage{arydshln}
\usepackage{setspace}
\usepackage{booktabs, pifont, xcolor}
\usepackage{colortbl}

\newcommand{\Ours}{\textsc{LongVideoAgent}\xspace}

\title{LongVideoAgent: Multi-Agent Reasoning with Long Videos}

  \author{
    Runtao Liu\thanks{~Equal Contribution. } \quad Ziyi Liu\footnotemark[1] \quad Jiaqi Tang \quad Yue Ma \quad Renjie Pi \quad Jipeng Zhang \quad Qifeng Chen \\
Hong Kong University of Science and Technology\\
{\tt\small rliuay@connect.ust.hk, ziyiliu0811@outlook.com}\\
\url{https://longvideoagent.github.io/}}

\begin{document}
\maketitle

\begin{abstract}
Recent advances in multimodal LLMs and systems that \emph{use tools} for long-video QA point to the promise of reasoning over hour-long episodes. 
However, many methods still compress content into lossy summaries or rely on limited toolsets, weakening temporal grounding and missing fine-grained cues. 
We propose a multi-agent framework in which a master LLM coordinates a grounding agent to localize question-relevant segments and a vision agent to extract targeted textual observations. 
The master agent plans with a step limit, and is trained with reinforcement learning to encourage concise, correct, and efficient multi-agent cooperation. 
This design helps the master agent focus on relevant clips via grounding, complements subtitles with visual detail, and yields interpretable trajectories. 
On our proposed \emph{LongTVQA} and \emph{LongTVQA+} which are episode-level datasets aggregated from TVQA/TVQA+, our multi-agent system significantly outperforms strong non-agent baselines. 
Experiments also show reinforcement learning further strengthens reasoning and planning for the trained agent. 

\end{abstract}

\section{Introduction}

Multimodal large language models (MLLMs) extend LLMs beyond text to perceive and reason over multimodal signals, such as visual frames, audio, and subtitles.
A key emerging challenge is robust \emph{long video} understanding, where information is sparsely distributed across hours of content and multiple modalities (e.g., frames, and dialogue cues).
Early instruction-tuned systems such as Video-LLaMA~\cite{zhang2023video,lin2024video} demonstrated that LLMs can be adapted to jointly process sampled video frames, marking an initial step toward multimodal video reasoning. However, current models remain limited to short clips or coarse summaries and struggle with fine-grained, temporally extended queries.
Crucially, most prior systems are \emph{non-agentic} models: they process a static, pre-encoded or down-sampled video. Converting the full visual stream into compressed representations in the LLM's textual space shifts the burden of temporal reasoning to this early stage—often lossy and irreversible, making it difficult to recover fine-grained evidence.
These limitations motivate an \emph{agentic}, tool-augmented paradigm that can actively decide what to observe next, when to query external visual or other tools, and when enough grounded evidence has been gathered to respond.
Despite recent advances, the field still lacks a solution that jointly achieves efficiency, multimodal completeness, and fine-grained temporal reasoning in long videos. 

\begin{figure}[t]
    \centering
    \includegraphics[width=0.48\textwidth]{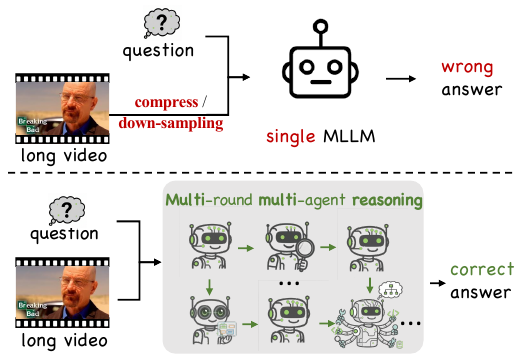}
    \caption{Traditional single-pass MLLMs that ingest entire long videos in one context—typically (may through heavy downsampling and compression) often miss crucial evidence and produce wrong answers, whereas \Ours{} conducts \emph{multi-agent}, \emph{multi-round}, and \emph{multimodal} reasoning to extract sparse, task-relevant cues and answer correctly.}
    \label{fig:lva_teaser}
    \vspace{-1em}
\end{figure}

Recent works have begun to frame long video understanding as an \emph{agent-driven process}, rather than a passive encoding task.
Notably, VideoAgent~\cite{fan2024videoagent,wang2024videoagent} introduced an agent-based framework where a central LLM actively conducts video analysis. In this paradigm, the LLM agent iteratively queries external vision models (tools) to retrieve and interpret video frames, progressively compiling the information needed to answer a given query. This interactive strategy mirrors human cognitive behavior and has demonstrated promising effectiveness. These findings highlight the potential of \emph{tool-augmented LLM agents} in achieving both efficiency and accuracy.
However, the initial incarnation of VideoAgent relies on a less powerful toolset, primarily generic vision-language foundation models for captioning and image retrieval. Such tools are often insufficient for capturing fine-grained semantics, precise object references, or subtle temporal cues. This restricts the agent’s ability to understand complex scenes and reason over long temporal spans.
Moreover, current frameworks underutilize the LLM’s inherent reasoning abilities and lack mechanisms for multi-step decision making or reinforcement-based planning.

In this paper, shown as Figure~\ref{fig:lva_teaser} we address these challenges by proposing a new \emph{multi-agent-based framework} for long video understanding that strategically incorporates agents.
Our system adopts a multi-agent architecture, where a central \textsc{MasterAgent} is responsible for reasoning and answering, while coordinating with other specialized agents.
Specifically, a \textsc{GroundingAgent} locates video segments relevant to the question, and a \textsc{VisionAgent} extracts detailed visual information from the selected clips (e.g., objects, faces, actions).
The master agent gathers these outputs to iteratively reason over the accumulated evidence.
To guide the reasoning process, we design a reward-driven training strategy that encourages the master agent to conduct structured, multi-step reasoning.
In each iteration, the master agent generates sub-queries, invokes either the grounding or vision agent as needed, and integrates the returned information before deciding on the next step.
When it determines that enough evidence has been collected, it produces a final answer.
By designing a reward function that penalizes irrelevant tool use and incoherent reasoning, we guide the agent to ``think'' in a proper format, effectively learning when to explore the video with tools and when it has gathered sufficient evidence to answer the question.
Furthermore, to evaluate long-form video reasoning in a realistic setting, we construct a new benchmark dataset \emph{LongTVQA} and \emph{LongTVQA+}.
This dataset extends the well-known TVQA video question answering task to much longer video durations, providing a rigorous testbed for our agent.

Our \emph{Agent-with-Tools} approach demonstrates superior performance on the LongTVQA benchmark, outperforming all existing baselines by a significant margin.
Through ablation studies, we show that both the multi-agent architecture and the reward-guided training contribute critically to the agent’s gains.
Our system not only achieves higher accuracy, but also exhibits interpretable decision-making, coordinating sub-agents to select relevant video segments and extract fine-grained visual information essential for reasoning. These results underscore the benefit of an agentic framework for long video understanding.

Our contributions are threefold: (i) a modular \emph{multi-agent} architecture in which a master LLM coordinates grounding and vision specialists; (ii) a \emph{reward-driven} agentic reinforcement learning training scheme that promotes concise, step-wise reasoning; and (iii) episode-level long video datasets LongTVQA and LongTVQA+ are proposed under which our system achieves state-of-the-art results.

\section{Related Work}
\subsection{Video Question Answering}
Early work focused on memory and attention mechanisms over appearance–motion features~\cite{gao2018comemory}. This evolved into multimodal transformers designed for efficient frame sampling~\cite{lei2021less}. 
Recent trends emphasize retrieval-aware reasoning and efficient tokenization for long videos, as well as integrating LLM-based reasoning with video encoders~\cite{zhang2023video} and employing agentic planners that iteratively gather evidence~\cite{wang2024videoagent}. 
Long-form systems further explore sparse memory and temporal grounding techniques to handle hour-scale inputs~\cite{song2024moviechat}. 
These developments motivate long-form VideoQA systems that selectively retrieve segments under a limited context budget.

\subsection{LLM Agents}
LLM agents couple chain-of-thought with \emph{actions}: planning, tool calls, and iterative evidence gathering. Foundational agent ideas include ReAct, Self-Ask, and WebGPT~\cite{yao2022react,press2022selfask,nakano2021webgpt}. 
Toolformer shows self-supervised API-calling, while orchestration frameworks (HuggingGPT/Gorilla-style) route subtasks to expert models~\cite{schick2023toolformer,shen2023hugginggpt}. 
In multimodal settings, MM-ReAct wires LLMs to vision experts via prompting, and program-of-thought systems like ViperGPT compose perception modules through executable code for transparent, verifiable reasoning~\cite{yang2023mmreact,suris2023vipergpt}. 
For long videos, agentic designs such as VideoAgent/VideoAgent-style frameworks use memory, targeted retrieval, and temporal grounding to operate under strict context budgets while improving faithfulness~\cite{wang2024videoagent}. 
Beyond planning, video-RAG pipelines extract ASR/OCR/objects and retrieve evidence to augment LVLMs for factual responses~\cite{luo2024video}. 
In addition, long-horizon multimodal agents with persistent memory and structural planning further enhance reliability for extended videos, e.g., Long-Seeing, VideoTree, and Koala~\cite{long2025seeing,wang2025videotree,tan2024koala}; and general reasoning paradigms such as Chain-of-Thought, Least-to-Most, Tree-of-Thoughts, and Generative Agents provide foundations for decomposition and memory~\cite{wei2022chain,zhou2022least,yao2023tree,park2023generative}. 
Retrieval-first paradigms like Retrieving-to-Answer complement agent pipelines with a retrieve-then-reason template~\cite{pan2023retrieving}. 
(We also include the alternative ReAct entry for key consistency~\cite{yao2022react}.)

\subsection{Multi-Modal LLMs}
Modern MLLMs combine strong vision encoders with instruction-tuned LLMs. CLIP pretraining provides broad visual–text transfer~\cite{radford2021clip}. 
Flamingo introduces a perceiver-style resampler for few-shot multimodal learning~\cite{alayrac2022flamingo}; BLIP-2/InstructBLIP bridge frozen encoders and LLMs~\cite{li2023blip2,dai2023instructblip}. 
Recent visually instruction-tuned MLLMs~\cite{tang2025robustr1degradationawarereasoningrobust, pi2024strengthening, pi2025pointing}, such as LLaVA~\cite{liu2023llava}, scale visual instruction tuning using open components, while LLaVA-OneVision~\cite{li2024llava} unifies high-resolution perception with token-efficient processing for both images and videos.
Recent video-tuned variants (e.g., Video-LLaVA) and training-free token schedulers (e.g., SlowFast-LLaVA) further improve temporal coverage and efficiency~\cite{lin2024video,xu2024slowfast}. 
Proprietary MLLMs (GPT-4/4o; Gemini 1.5) show long-context multimodal reasoning~\cite{achiam2023gpt4,gemini2024}, while open models (Qwen2-VL, InternVL) narrow the gap via dynamic resolution, OCR, and video pipelines~\cite{Wang2024Qwen2VLEV,chen2024fargpt4vclosinggap}. 
Complementary advances focus on unifying image–video tokens with few, informative representations (e.g., MiniGPT4-Video, Video-ChatGPT, Video-LaVIT, LLaMA-VID, LongVU, PLLaVA, LLaVA-Video, Chat-UniVi)~\cite{ataallah2024minigpt4,maaz2024video,jin2024video,li2024llama,shen2024longvu,xu2024pllava,zhang2024video,jin2024chat}, 
and on long-context optimization or adaptive input selection (e.g., InternVideo2.5, LongVLM, Long Context training, self-adaptive sampling, simple-but-effective alignment, and question-instructed tuning)~\cite{wang2025internvideo2,weng2024longvlm,zhang2024long,han2023self,zhang2024simple,romero2024question}. 
Comprehensive analyses of video understanding in large multimodal models (e.g., Apollo) situate these models within broader capabilities and evaluation protocols~\cite{zohar2025apollo}. 
For key harmonization with the bibliography, we also include the alternate Video-LLaMA entry~\cite{zhang2023video}. 
However, most models still face long-video constraints (context length, retrieval). This motivates combining video-native encoders, instruction tuning, retrieval, and tool use for scalable long-form VideoQA.

\section{Method}
\label{sec:method}

\begin{figure*}[t]
  \centering
  \includegraphics[width=\textwidth]{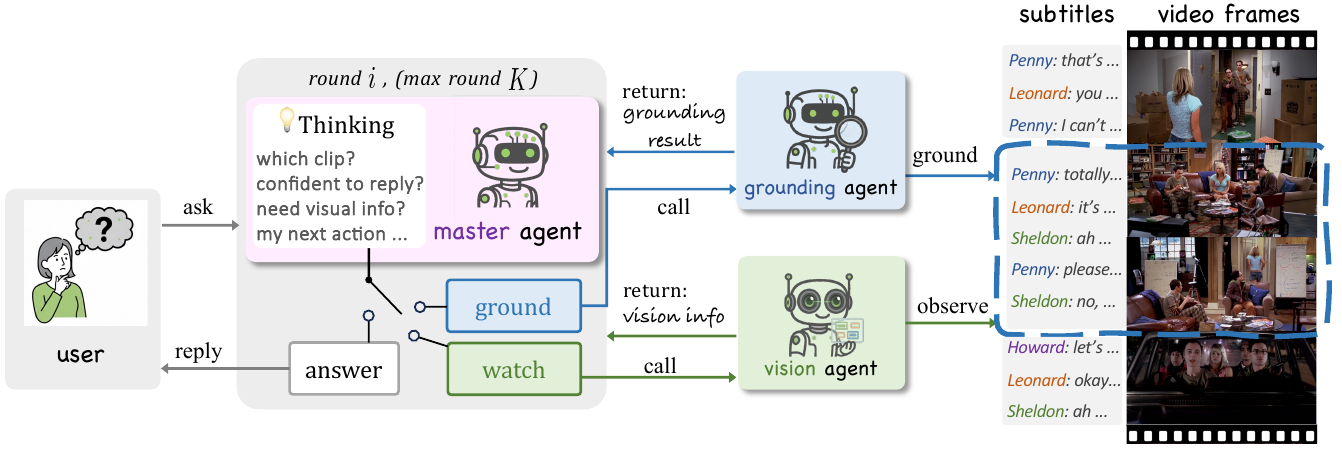}
  \vspace{-20pt}
  \caption{Architecture of \Ours{}. A \textsc{MasterAgent} runs for up to $K$ rounds, collaborating with a \textsc{GroundingAgent}  to localize relevant clips from videos and a \textsc{VisionAgent} to read fine-grained cues from the localized frames. Evidence accumulates until the \textsc{MasterAgent} feels confident to answer the user.}
  \label{fig:arch}
\end{figure*}

\begin{table*}[h]
\caption{System prompt for \Ours{}.}
\vspace{-5pt}
\label{tab:system-prompt}
\centering
\renewcommand{\arraystretch}{1.12}
\begin{spacing}{0.9} %
\small
\begin{tabular}{p{0.96\linewidth}}
\toprule
\textbf{System Prompt — \Ours{}}\\
\hline
You are an agent that answers questions about a long video episode. You may use two tools: a \emph{grounding agent} to localize relevant segments and a \emph{vision agent} to extract visual facts from the localized segment. Produce \emph{concise, direct} answers.\\[0.35em]
\textbf{Context you may receive.} All subtitles and the user question $q$. When a segment has been localized, you will also have a tag \texttt{\textless clipX\textgreater} (e.g., \texttt{\textless clip2\textgreater}). When the vision agent has been called, you will see its textual response.\\[0.35em]
\textbf{Available actions (choose exactly one per turn).}\\
\textbf{A — Visual query:} If current visual information is insufficient, or you need visual details conditioned on the subtitles for the current \texttt{\textless clipX\textgreater}, call the vision engine with \texttt{\textless visual\_query\textgreater query \textless /visual\_query\textgreater}.\\
\textbf{B — (Re)Grounding:} If the current text/visual evidence conflicts with the question, or the current location cannot support a confident answer, call the grounding agent with \texttt{\textless request\_grounding\textgreater}.\\
\textbf{C — Answer:} If evidence is sufficient, return the final answer with \texttt{\textless answer\textgreater ... \textless /answer\textgreater}. The answer must be concise and direct.\\[0.25em]
\textbf{Guidelines.} (1) Be conservative with tool calls; answer when sufficient. (2) Do not hallucinate visual details; only use the vision agent for facts not inferable from subtitles. (3) Each turn targets the current \texttt{\textless clipX\textgreater} (if any); if none exists, prefer (re)grounding before visual query.\\
\bottomrule
\end{tabular}
\end{spacing}
\end{table*}

As shown in Figure~\ref{fig:arch}, we cast long-video QA as \emph{multi-agent reasoning}, where a master agent LLM coordinates a grounding agent to temporally localize question-relevant segments and a vision agent to extract targeted observations from those segments. 
The system proceeds iteratively, maintaining a running context that accumulates subtitles, relevant segment tags, and vision observations, and it produces an answer once the master agent judges that sufficient evidence has been gathered. 
For open-source LLMs serving as the master agent, we apply reinforcement learning to encourage accurate, concise, and cooperation-efficient behavior while keeping the other agents frozen.
At inference, the process yields clear, step-by-step traces aimed at solving the question at hand.

\subsection{Multi-agent System Framework}

\paragraph{Master agent behavior and training.}
Specifically, the master agent follows the instruction schema in the \emph{System Prompt} (Table~\ref{tab:system-prompt}) and the multi-turn policy in Algorithm~\ref{alg:long_video_agent} that coordinates two other specialist agents: a grounding agent and a vision agent. 
Given an episode with its full subtitles and a question, the master runs a bounded loop (at most $K$ steps). At each turn it emits exactly one structured action token, \verb|<watch>| for a visual read, \verb|<request_grounding>| for (re)localization, or \verb|<answer>| to terminate. After the corresponding agent is invoked, its textual output is appended to the context of the master agent. For open-source masters, we optimize the policy with GRPO while keeping the grounding and vision agents fixed. The rollouts terminated by action tokens in Algorithm~\ref{alg:long_video_agent} provide the trajectories for training and evaluation.

\paragraph{Grounding agent.}
Given the question and subtitles, the grounding agent proposes a temporal segment and returns a symbolic tag \verb|<clip_X>| marking the relevant portion of the episode. By default the window context is $1$; when larger, the agent outputs a short run of consecutive tags. The master may re-query grounding to refine or validate the segment as reasoning progresses.

\paragraph{Vision agent.}
Conditioned on \verb|<clip_X>| and an on-demand prompt that specifies the current visual need, the vision agent extracts textual observations from frames within the localized segment (e.g., objects/entities, attributes, actions, OCR/on-screen text, scene cues). These observations are appended to the context and guide the next decision; the loop terminates when the master judges the accumulated visual evidence sufficient to answer.

\begin{algorithm*}[h]
\caption{\Ours{} with Multi-Turn Reasoning}
\label{alg:long_video_agent}
\begin{spacing}{0.95}       %
\small                      %
\begin{algorithmic}[1]
\REQUIRE Subtitles $\mathcal{S}$; question $q$; video $V$; \textsc{MasterAgent} parameters $\pi_{\theta}$; maximum steps $K$; \textsc{GroundingAgent}; \textsc{VisionAgent}.
\ENSURE Final answer $\hat{y}$.

\STATE Initialize rollout sequence $y \gets \emptyset$
\STATE Initialize step count $t \gets 0$

\WHILE{$t < K$}
    \STATE Initialize current action LLM rollout sequence $y_t \gets \emptyset$
    \WHILE{True}
        \STATE Generate thinking token $y_i \sim \pi_{\theta}(\cdot \mid \mathcal{S}, q, V, y + y_t)$
        \STATE Append $y_i$ to rollout sequence $y_t \gets y_t + y_i$
    \IF{$y_i$ in [\texttt{</visual\_query>}, \texttt{</request\_grounding>}, \texttt{</answer>}, \texttt{<eos>}]}
            \STATE \textbf{break}
        \ENDIF
    \ENDWHILE
    
    \STATE $y \gets y + y_t$
    
    \IF{\texttt{<visual\_query>} detected in $y_t$}
        \STATE Extract visual query $q_{vis} \gets \textsc{Parse}(y_t, \texttt{<visual\_query>}, \texttt{</visual\_query>})$
        \STATE Retrieve vision results $d = \textsc{VisionAgent}(q_{vis}, V)$
        \STATE Insert visual results into rollout $y \gets y + d$
    \ELSIF{\texttt{<request\_grounding>} detected in $y_t$}
        \STATE Retrieve grounding results $\texttt{clipTag} = \textsc{GroundingAgent}(q, \mathcal{S})$
        \STATE Insert clip tag into rollout $y \gets y + \texttt{clipTag}+\mathcal{S}(\texttt{clipTag})$
    \ELSIF{\texttt{<answer>} detected in $y_t$}
        \STATE Extract predicted answer $\hat{y} \gets \textsc{Parse}(y_t,\texttt{<answer>},\texttt{</answer>})$; Normalize $\hat{y}$ (trim spaces/punctuation)
        \STATE Insert final answer into rollout $y \gets y + \text{"The answer is: "}+ \hat{y}$
        \STATE \textbf{return} final answer $\hat{y}$
    \ELSE
        \STATE Ask for rethink $y \gets y+\text{"The action is not correct. Only <visual\_query>, <request\_grounding>, or <answer>."}$
    \ENDIF
    
    \STATE Increment step count $t \gets t + 1$
\ENDWHILE

\STATE \textbf{return} final generated response $y$ for $q$
\end{algorithmic}
\normalsize
\end{spacing}
\end{algorithm*}

\subsection{Reinforcement Learning for \textsc{LongVideoAgent}}
For open-source LLMs serving as the master agent, we fine-tune the master with GRPO while keeping the grounding and vision agents frozen. 
Long-video QA is cast as a finite-horizon decision process: at each action step after reasoning the policy emits exactly one structured action token (\verb|<visual\_query>|, \verb|<request_grounding>|, or \verb|<answer>|). 

\paragraph{Trajectory.}
A full response terminates upon emitting \verb|<answer>...</answer>| or reaching $K$ steps. We index decision steps by $t\in\{0,1,\ldots,T\}$ with $T\le K$. At step $t$, the policy $\pi_{\theta}$ first plans and then emits a contiguous action string $a_t$ ending with exactly one closing tag from \{\texttt{</visual\_query>}, \texttt{</request\_grounding>}, \texttt{</answer>}\}. 
If not terminating, the system appends feedback from the invoked agent $o_t$ (e.g., a vision observation or a clip tag) to the context for the next step. 

\paragraph{Rewards.}
We use two simple, rule-based rewards as supervision for reinforcement learning:
(i) \emph{Structural validity} $r^{\text{fmt}}_t\in\{0,1\}$ grants $1$ if the action string contains exactly one top-level tag with proper closure and no extraneous text; otherwise $0$.
(ii) \emph{Answer correctness} $r^{\text{ans}}\in[0,1]$ is awarded at termination via exact match on the multiple-choice answer; if no valid \verb|<answer>| appears, $r^{\text{ans}}=0$.

\paragraph{Objective and optimization.}
We seek a policy that produces well-formed actions at every step and a correct final answer. To balance these goals, the trajectory reward return is
$R(\tau)=\alpha \sum_{t=0}^{T} r^{\text{fmt}}_t + r^{\text{ans}}$
where $\alpha>0$ weights the per-step structural shaping and $r^{\text{ans}}$ supplies the terminal task reward. $r^{\text{fmt}}_t$ encourages the master to emit exactly one correct action tag at each decision, while $r^{\text{ans}}$ evaluates only the final \verb|<answer>|. If no valid and correct answer is produced, $r^{\text{ans}}=0$. 

We optimize the master agent with GRPO on sampled rollouts: for each episode, the policy generates an action sequence, receives structural rewards at action boundaries and a terminal answer reward, and we compute sequence-level advantages with a learned value baseline. 
Policy updates follow the GRPO objective with standard clipping and entropy regularization, while the grounding and vision agents remain frozen. 
This minimal, two-signal objective provides sufficient guidance to learn structured, multi-turn coordination without additional dense rewards. 

\section{Experiments}
\label{sec:main-results}

\newcommand{\cmark}{\ding{51}} %
\newcommand{\xmark}{\ding{55}} %
\newcommand{\gcmark}{\textcolor{green!55!black}{\ding{51}}}

\newcommand{\gain}[1]{(\textcolor{green!60!black}{+#1})}

\begin{table*}[ht]
\centering
\footnotesize
\caption{Performance on \emph{LongTVQA} and \emph{LongTVQA+}. 
The left block lists model attributes (\emph{Agentic}, \emph{Input}, \emph{RL fine-tune}); 
the right block reports validation accuracy (\%). 
GPT-4o and Gemini-2.5 Pro are \emph{multimodal} baselines that process and accept the full long video directly. 
Methods labeled \texttt{Agentic} indicate the model operates as the \textsc{MasterAgent}; 
methods labeled \texttt{AgenticRL} additionally denote RL fine-tuning.
Parenthesized \textcolor{green!60!black}{green} numbers denote absolute gains over the immediately preceding (non-agentic or non-RL) setting.
We observe that: (i) our multi-agent framework, \Ours{}, consistently outperforms the non-agentic counterparts; 
(ii) agentic RL yields additional gains, especially for smaller open-source models; 
(iii) using frames provides visual evidence beyond subtitles, and generally outperforms subtitle-only inputs; 
(iv) closed-source models remain strong, but the gap narrows much when open-source models adopt agentic designs and agentic RL.}

\label{tab:baseline-landscape}
\begin{tabular}{lcccll}
\toprule
\textbf{\multirow{2}{*}{Method}} & 
\textbf{\multirow{2}{*}{Multi-agent}} & 
\textbf{\multirow{2}{*}{Input}} & 
\textbf{\multirow{2}{*}{RL Finetune}} & 
\multicolumn{2}{c}{\textbf{Accuracy (\%)}} \\
\cmidrule(lr){5-6}
 &  &  &  & LongTVQA & LongTVQA+ \\
\midrule
\multicolumn{6}{l}{\textcolor{gray}{\emph{Closed-source (M)LLMs}}} \\
GPT-4o           & \xmark & Subtitle+Frame   & \xmark & 70.78 & 78.32 \\
Gemini-2.5 Pro   & \xmark & Subtitle+Frame   & \xmark &  \underline{78.90}   & \underline{81.28} \\
\hdashline
GPT5-mini        & \xmark & Subtitle         & \xmark & 62.40 & 66.70 \\
Agentic-GPT5-mini        & \gcmark & Subtitle+Frame   & \xmark & 71.11\gain{8.71} & 78.90\gain{12.20} \\
Grok             & \xmark & Subtitle         & \xmark & 76.90 & 81.80 \\
Agentic-Grok             & \gcmark & Subtitle+Frame   & \xmark & \textbf{82.65}\gain{5.75} & \textbf{85.60}\gain{3.80} \\
\midrule
\multicolumn{6}{l}{\textcolor{gray}{\emph{Open-source LLMs}}} \\
DeepSeek-R1(671B)         & \xmark & Subtitle         & \xmark & 68.99 & 75.04 \\
Agentic-DeepSeek-R1(671B)         & \gcmark & Subtitle+Frame   & \xmark & 70.30\gain{1.31} & 79.70\gain{4.66} \\
\hdashline
Agentic-Qwen2.5(3B)       & \cmark & Subtitle+Frame   & \xmark & 23.50 & 27.70 \\
AgenticRL-Qwen2.5(3B)   & \cmark & Subtitle+Frame & \gcmark & 47.40\gain{23.90} & 50.10\gain{22.40} \\
Agentic-Qwen2.5(7B)       & \cmark & Subtitle+Frame   & \xmark & 46.10 & 60.30 \\
AgenticRL-Qwen2.5(7B)       & \cmark & Subtitle+Frame   & \gcmark & \textbf{60.20}\gain{14.10} & \textbf{70.80}\gain{10.50} \\
\bottomrule
\vspace{-20pt}
\end{tabular}
\end{table*}

\begin{table}[h]
    \centering
    \begin{spacing}{0.9}
    \footnotesize
    \caption{A case of \Ours{} showing the collaboration among multiple agents.}
    \label{tab:case-sheldon}
    \begin{tabular}{p{0.96\linewidth}}
        \toprule
        \textbf{Question \& Choices} \\
        \midrule
        Question: Where is Sheldon sitting when he is accompanied by a man? \\
        Choices: a0: The Mall \quad a1: A Movie Theatre \quad a2: A Park \quad \textbf{a3: A Bus Stop}~\textcolor{green}{\checkmark} \quad a4: The Store \\
        \midrule
        \textbf{\Ours{} Execution} \\
        \midrule
        \textbf{\textsc{Master Agent}}: \\
        \texttt{\textcolor{blue}{<think>}} Subtitles alone do not specify the location; I will first localize with grounding, then request a visual read. \texttt{\textcolor{blue}{</think>}} \\
        \texttt{\textcolor{purple}{<request\_grounding>}} $\Rightarrow$ \textsc{Grounding Agent} \\
        \hdashline
        \textbf{\textsc{Grounding Agent}}: \\
        \textit{Grounding information.} Subtitles: \emph{“Sheldon: … apparently we’re just two peas in a pod. … Mrs Cooper: Here. Thank you.”} \quad
        Clip tag: \texttt{<s05e06\_seg02\_clip\_15>} \\
        \hdashline
        \textbf{\textsc{Master Agent}}: \\
        \texttt{\textcolor{blue}{<think>}} The grounded subtitles still don’t reveal the setting; I’ll request a visual description of the localized clip. \texttt{\textcolor{blue}{</think>}} \\
        \texttt{\textcolor{purple}{<visual\_query>}} visual description of scene in \emph{The Big Bang Theory} S05E06 segment 02 clip 15 where Sheldon is with a man \texttt{\textcolor{purple}{</visual\_query>}} $\Rightarrow$ \textsc{Vision Agent} \\
        \hdashline
        \textbf{\textsc{Vision Agent}}: \\
        \textit{Visual description.} Sheldon is sitting on a bench at night in an urban sidewalk setting; a waste container and building windows are visible nearby. \\
        \hdashline
        \textbf{\textsc{Master Agent}}: \\
        \texttt{\textcolor{blue}{<think>}} Bench + sidewalk + trash can + windows strongly indicate a bus stop rather than a mall, theatre, store, or park. No further tools needed. \texttt{\textcolor{blue}{</think>}} \\
        \texttt{\textcolor{purple}{<answer>}} a3: A Bus Stop~\textcolor{green}{\checkmark} \texttt{\textcolor{purple}{</answer>}} $\Rightarrow$ \textsc{User} \\
        \bottomrule
    \end{tabular}
\normalsize
\end{spacing}
\end{table}

\subsection{Datasets}
\label{sec:datasets}
We build \emph{LongTVQA} and \emph{LongTVQA+} on top of TVQA and TVQA+. 
TVQA spans six TV shows with 152.5K multiple-choice QAs over 21.8K clips (60–90s) with subtitles and moment annotations; questions require joint dialogue–visual reasoning~\cite{lei2018tvqa}. TVQA+ refines a subset with spatio-temporal grounding—adding precise timestamps and 310.8K frame-level boxes for referenced entities (29.4K QAs from 4{,}198 clips, mainly TBBT)—supporting joint QA and temporal/spatial localization~\cite{lei2020tvqa+}.

To obtain \emph{LongTVQA} and \emph{LongTVQA+}, we aggregate all clips from the same TV episode into a single \emph{episode-level} (hour-scale) sequence. For each episode, we merge the visual stream, subtitles, and all associated questions; clip timestamps are re-indexed into the episode timeline, and TVQA+ bounding boxes are preserved at their corresponding frames. Unless otherwise noted, we report results on the original validation splits after this episode-level aggregation.

\subsection{Baselines}
\label{sec:baselines}
We include both open-source and closed-source models (see Table~\ref{tab:baseline-landscape}), including representative open-source LLMs such as \emph{DeepSeek-R1}~\cite{guo2025deepseek} and \emph{Qwen2.5-3B/7B} ~\cite{qwen2025qwen25technicalreport} , and closed-source models such as \emph{Grok}, \emph{GPT5-mini}~\cite{gpt5}, GPT-4o ~\cite{gpt4o} and \emph{Gemini 2.5 Pro}~\cite{comanici2025gemini}. We adopt unified experimental settings to ensure comparability across backbones.
We evaluate \emph{base LLMs in a non-agent mode} and \emph{our agent system}, with the base LLM serving as the master agent in both cases. In the non-agent mode, the model consumes the full subtitles and does not invoke grounding or vision agents. We then compare our agent system against the corresponding non-agent runs on the \emph{same} backbones, so that observed gains can be attributed to agentic behavior rather than backbone differences.
For open-source backbones, we also report results \emph{with vs.\ without} reinforcement learning under the identical evaluation protocol described in \S\ref{sec:setup}. Closed-source models are evaluated as released, without additional training.

\subsection{Experimental Setup}
\label{sec:setup}
By default we use Grok-4-fast-reasoning for temporal localization and GPT\mbox{-}4o as the vision agent. 
The window context is set to $1$, meaning the agent conditions on a single localized clip (no adjacent clips), and the maximum execution steps are $K{=}5$. 
All methods read the full episode subtitles. In the non-agent setting, no external modules are invoked. 
In the agent setting, the master agent receives symbolic temporal tag(s) \texttt{<clip\_X>}, which is produced by the grounding agent, that marks the grounded clip(s) on the episode timeline, and it may request on-demand \emph{textual} observations from the vision agent via prompting for selected frames within that segment (e.g., objects and attributes, OCR/text, brief scene cues). 
The master agent only consumes text (subtitles, the \texttt{<clip\_X>} tag, and optional visual observations); no raw images are passed to the master agent. 
We report results on the \emph{validation} splits of LongTVQA and LongTVQA+, using \emph{Answer Accuracy (Acc)} as the primary metric (the questions are multiple choice) and additionally \emph{Grounding Accuracy} for experiments that involve clip grounding. 
For reinforcement learning, we use GRPO with a learning rate of $5\times10^{-6}$, up to 2{,}000 optimization steps, a KL coefficient of $10^{-3}$, batch size $4$, rollout count $N{=}4$, and temperature $1.0$. 
Training Qwen2.5-7B took ~12 hours on 4× NVIDIA H800 GPUs, while the 3B variant took ~6 hours under the same setup.

\subsection{Performance}
\label{sec:perf}
Table~\ref{tab:baseline-landscape} presents overall validation accuracy. Moving from the non-agent setting to our multi-agent framework yields significante gains. This provides direct evidence for the effectiveness of a \emph{multi-agentic} pipeline that can localize the relevant clips and performs targeted visual inspection. In addition, for several open-source LLMs(as master agent), reinforcement learning consistently improves over their inference-only counterparts under identical prompts and evaluation; notably, the Qwen2.5-7B model with RL attains accuracy comparable to \textit{GPT-5-mini} (closed-source) on our protocol. Illustrative examples in Table~\ref{tab:case-sheldon} and Table~\ref{tab:case-bedside-multi} effectively demonstrate the efficacy of our approach, with additional cases provided in the supplementary materials.

\begin{table}[t]
\centering
\small
\caption{Ablations and analysis of \Ours{}.}
\label{tab:all-results}
\begin{subtable}[t]{0.48\textwidth}
\centering
\caption{\textbf{Comparison of non-agent vs. multi-agent performance.} Agentic components progressively improve performance: adding grounding outperforms the non-agent baseline, and adding vision agent yields the best results.}
\label{tab:main-results}
\begin{tabular}{lc}
\toprule
Setting & \textbf{Accuracy (\%)} \\
\midrule
Non-agent (Text-only)      & 64.3 \\
Multi-Agent (Grounding)          & 69.0 \\
Multi-Agent (Grounding + Vision) & \textbf{74.8} \\
\bottomrule
\end{tabular}
\end{subtable}\hfill
\begin{subtable}[t]{0.48\textwidth}
\centering
\caption{\textbf{Effect of max steps $K$.} Increasing the \textsc{MasterAgent} step budget generally raises both grounding and overall accuracy until reaching a saturation position.}
\label{tab:ablate-steps}
\begin{tabular}{lcc}
\toprule
$K$ & \textbf{Grounding Accuracy (\%)} & \textbf{Accuracy (\%)} \\
\midrule
2  & 67.00 & 68.30 \\
5  & 71.00 & 73.67 \\
10 & 72.00 & 73.67 \\
\bottomrule
\end{tabular}
\end{subtable}

\vspace{0.6em}

\begin{subtable}[t]{0.48\textwidth}
\centering
\caption{\textbf{Effect of evidence window size.} Larger temporal windows supply richer context for grounding and vision. }
\label{tab:ablate-window}
\begin{tabular}{lcc}
\toprule
Window & \textbf{Grounding Accuracy (\%)} & \textbf{Accuracy (\%)} \\
\midrule
1 & 71.67 & 70.33 \\
2 & 78.67 & 75.00 \\
3 & \textbf{81.94} & \textbf{77.26} \\
\bottomrule
\end{tabular}
\end{subtable}\hfill
\begin{subtable}[t]{0.48\textwidth}
\centering
\caption{\textbf{Vision model ablation.} Stronger \textsc{VisionAgent} shows higher overall accuracy, reflecting higher quality extraction of visual information from frames.}
\label{tab:ablate-vision}
\begin{tabular}{lcc}
\toprule
Vision model & \textbf{Grounding Acc. (\%)} & \textbf{Acc. (\%)} \\
\midrule
Qwen3-VL-235B & 71.00 & 73.67 \\
GPT-4o             & 73.30 & \textbf{78.00} \\
\bottomrule
\end{tabular}
\end{subtable}
\end{table}

\subsection{Ablation Studies and Analysis}

\paragraph{Execution step limit $K$.}
Table~\ref{tab:ablate-steps} varies the \emph{upper bound} on agent actions per question. Increasing $K$ from $2$ to $5$ raises temporal localization accuracy from $67.0$ to $71.0$ (\,+4.0\,) and answer accuracy from $68.30$ to $73.67$ (\,+5.37\,). Raising the limit to $K{=}10$ yields no additional gain in answer accuracy ($73.67$) while localization nudges to $72.0$, indicating diminishing returns. We therefore adopt $K{=}5$ as the default.

\paragraph{Window context.}
With GPT\mbox{-}4o as the vision agent, expanding the evidence window from $1$ to $2$ \emph{adjacent clips} improves localization from $71.67$ to $78.67$ (\,+7.00\,) and answer accuracy from $70.33$ to $75.00$ (\,+4.67\,); using $3$ clips yields further gains to $81.94$ and $77.26$ (additional +3.27 and +2.26), as shown in Table~\ref{tab:ablate-window}. The adjacent context helps disambiguate references across shots, but larger windows require more visual queries and latency while delivering smaller accuracy gains; hence we keep \emph{Window}$=1$ for the default and study larger windows in ablations.

\begin{table}[ht]
    \centering
    \begin{spacing}{0.95}
    \footnotesize
    \caption{Example showing \Ours{} reasoning and repeatedly invoking the \textsc{Grounding} and \textsc{Vision} agents to gather evidence and answer correctly.}
    \label{tab:case-bedside-multi}
    \begin{tabular}{p{0.96\linewidth}}
        \toprule
        \textbf{Question \& Choices} \\
        \midrule
        Question: What side of the bed is Sheldon when he is closer to the window? \\
        Choices: \textbf{a0: the left side}~\textcolor{green}{\checkmark} \quad a1: the right side \quad a2: the middle \quad a3: not on the bed \quad a4: the chair \\
        \midrule
        \textbf{\Ours{} Execution} \\
        \midrule
        \textbf{\textsc{Master Agent}}: \\
        \texttt{\textcolor{blue}{<think>}} Subtitles alone cannot disambiguate left vs.\ right; I will first ground the segment, then request visual evidence. \texttt{\textcolor{blue}{</think>}} \\
        \texttt{\textcolor{purple}{<request\_grounding>}} localize the referenced segment \texttt{\textcolor{purple}{</request\_grounding>}} $\Rightarrow$ \textsc{Grounding Agent} \\
        \hdashline
        \textbf{\textsc{Grounding Agent}}: \\
        \textit{Grounding information.} Subtitles (localized): \texttt{<s10e04\_seg02\_clip\_09>} \emph{“Sheldon: Okay. Now, on this side, … Bernadette: Raj knows,”} \\
        \hdashline
        \textbf{\textsc{Master Agent}}: \\
        \texttt{\textcolor{blue}{<think>}} Grounded subtitles still don’t reveal which side is nearer the window; I’ll request a visual read. \texttt{\textcolor{blue}{</think>}} \\
        \texttt{\textcolor{purple}{<visual\_query>}} Big Bang Theory season 10 episode 4 — bedroom scene layout (Sheldon \& Amy); identify bed/window configuration \texttt{\textcolor{purple}{</visual\_query>}} $\Rightarrow$ \textsc{Vision Agent} \\
        \hdashline
        \textbf{\textsc{Vision Agent}}: \\
        \textit{Visual description (1).} Bedroom scene; bed centered against the wall; Sheldon sits on the bed; general layout and decor described, but the side nearest the window is not specified. \\
        \hdashline
        \textbf{\textsc{Master Agent}}: \\
        \texttt{\textcolor{blue}{<think>}} The first visual read lacks the window–bed side relation; I need a more precise visual query focused on the window’s position relative to Sheldon. \texttt{\textcolor{blue}{</think>}} \\
        \texttt{\textcolor{purple}{<visual\_query>}} The Big Bang Theory S10E04 seg02 clip 09 — which side of the bed is next to the window, and where is Sheldon relative to it? \texttt{\textcolor{purple}{</visual\_query>}} $\Rightarrow$ \textsc{Vision Agent} \\
        \hdashline
        \textbf{\textsc{Vision Agent}}: \\
        \textit{Visual description (2).} A window with patterned curtains is on the \emph{left} side of the bed; Sheldon is seated against the headboard near that window. \\
        \hdashline
        \textbf{\textsc{Master Agent}}: \\
        \texttt{\textcolor{blue}{<think>}} Window is on the left side of the bed and Sheldon is closer to it; therefore he is on the left side. No further tools needed. \texttt{\textcolor{blue}{</think>}} \\
        \texttt{\textcolor{purple}{<answer>}} a0: the left side~\textcolor{green}{\checkmark} \texttt{\textcolor{purple}{</answer>}} $\Rightarrow$ \textsc{User} \\
        \bottomrule
    \end{tabular}
\normalsize
\end{spacing}
\end{table}

\paragraph{Vision model.}
Table~\ref{tab:ablate-vision} compares perception backbones. GPT\mbox{-}4o attains $73.30$ localization and $78.00$ answer accuracy, outperforming Qwen3\mbox{-}VL\mbox{-}235B\mbox{-}a22b at $71.00$ and $73.67$ by +2.30 and +4.33, respectively. The gap indicates that stronger visual recognition (small objects, OCR, fine attributes) translates into better end-task accuracy in long-form QA, so we adopt GPT\mbox{-}4o as the default vision agent.

\paragraph{Contribution of agentic components.}
Table~\ref{tab:main-results} decomposes the gains when moving \emph{from a single LLM} to a \emph{multi-agent, multimodal} system. Adding temporal grounding to the same backbone increases answer accuracy from $64.3$ to $69.0$ (+4.7), showing that identifying the relevant clip filters distractors and focuses reasoning. Enabling vision after grounding further lifts accuracy to $74.8$ (+5.8 over grounding; +10.5 overall): targeted visual inspection complements subtitles with concrete object/text cues and can validate or refine grounding through repeated calls when uncertain. Because backbones and prompts are held fixed, these improvements are attributable to the agentic procedure. We suggest grounding narrows the context length for reasoning and guides the master agent's attention, while vision supplies the missing fine-grained evidence.

\section{Conclusion}
We presented a multi-agent framework, \Ours{}, for long-form video question answering in which a \textsc{Master} agent coordinates a \textsc{GroundingAgent} for temporal localization and a \textsc{VisionAgent} for targeted perception. 
The framework is model-agnostic: we evaluate it with both closed- and open-source LLMs; for open-source masters, we fine-tune with GRPO to encourage accurate, concise, and cooperation-efficient behavior while keeping the other agents frozen. 
Equipped with a unified context and GRPO training that combines structural and answer rewards, the system where open-source LLMs act as the master agent yields transparent, step-by-step traces and achieves strong gains on \emph{LongTVQA} / \emph{LongTVQA+} over non-agent baselines. 
Ablations show that grounding+vision is essential, modest step limits suffice, adjacent-window context helps, and stronger perception yields higher accuracy, validating the effectiveness of the framework. Future work includes richer modalities(like audio track and knowledge background), finer grounding and larger-scale RL training.

\clearpage

\section*{Limitations}
Our work has several practical limitations. 
First, based on TVQA and TVQA+, we rely on provided subtitles as the primary textual channel and do not process raw audio; in future work we plan to integrate an audio-to-subtitles (ASR) module to capture raw speech. 
Second, the vision and grounding modules are kept fixed during RL. Jointly optimizing them could further improve robustness and accuracy. 
Lastly, the reward is intentionally simple (format + answer correctness), which may still have room for improvements.

\bibliography{custom}
\clearpage

\end{document}